\documentclass[letterpaper, 10 pt, conference]{ieeeconf}  

\IEEEoverridecommandlockouts                              

\usepackage{graphics} 
\usepackage{epsfig} 
\usepackage{mathptmx} 
\usepackage{times} 
\usepackage{amsmath} 
\usepackage{amssymb}  

\usepackage{eurosym}
\usepackage{cite}
\usepackage{textcomp}
\usepackage{xspace}
\usepackage{xcolor} 
\usepackage{censor}
\usepackage{multirow}
\usepackage{booktabs} 
\usepackage{array}
\usepackage{placeins}
\usepackage{flushend}
\usepackage{hyperref}

\usepackage{blindtext}

\usepackage[font=small,labelfont=bf]{caption}

\title{\LARGE \bf
Real-time Rendering-based Surgical Instrument Tracking via Evolutionary Optimization
}

\author{
Hanyang Hu$^{1}$, Zekai Liang$^{1}$, Florian Richter$^{1}$, Michael C. Yip$^{1}$, \textit{Senior Member, IEEE}
\thanks{$^{1}$Department of Electrical and Computer Engineering, University of California San Diego, La Jolla, CA 92093 USA.
        {\tt\small \{hanyang-hu, z9liang, frichter, yip\}@ucsd.edu}}%
}

\DeclareMathAlphabet{\mathcal}{OMS}{cmsy}{m}{n}
\DeclareMathOperator*{\argmax}{arg\,max}
\DeclareMathOperator*{\argmin}{arg\,min}

\begin{document}

\maketitle
\thispagestyle{empty}
\pagestyle{empty}

\begin{abstract}

Accurate and efficient tracking of surgical instruments is fundamental for Robot-Assisted Minimally Invasive Surgery. Although vision-based robot pose estimation has enabled markerless calibration without tedious physical setups, reliable tool tracking for surgical robots still remains challenging due to partial visibility and specialized articulation design of surgical instruments. Previous works in the field are usually prone to unreliable feature detections under degraded visual quality and data scarcity, whereas rendering-based methods often struggle with computational costs and suboptimal convergence. In this work, we incorporate CMA-ES, an evolutionary optimization strategy, into a versatile tracking pipeline that jointly estimates surgical instrument pose and joint configurations. Using batch rendering to efficiently evaluate multiple pose candidates in parallel, the method significantly reduces inference time and improves convergence robustness. The proposed framework further generalizes to joint angle-free and bi-manual tracking settings, making it suitable for both vision feedback control and online surgery video calibration. Extensive experiments on synthetic and real-world datasets demonstrate that the proposed method significantly outperforms prior approaches in both accuracy and runtime. Source code and data are available at \url{https://github.com/hanyang-hu/online_dvrk_tracking}.

\end{abstract}

\section{INTRODUCTION}

Robot-assisted Minimally-Invasive Surgery (RMIS) has been increasingly adopted to improve surgical safety and procedural efficiency in complex, long-duration operations under physically demanding conditions. Accurate and efficient pose estimation of articulated surgical instruments
 is essential for a wide range of downstream applications, such as vision-based control \cite{9217086}, augmented reality guidance \cite{Kalia2021}, and robot learning \cite{kim2024surgicalrobottransformersrt}.

 
Uneven geometric structures, partial visibility of the kinematic chain, and noisy joint angle measurements caused by cable stretch introduce additional challenges for accurate articulated instrument pose recovery. Previous works have approached this problem from both kinematic calibration and pose calibration perspectives. 
Early works such as \cite{zhang2013optical, jun2018using, 8460583, 9312948, ozguner2020camera} proposed external sensing, including optical tracking and depth/vision systems, to compensate for kinematic error. However, these approaches require complex physical setups and are difficult to replicate or deploy in new environments. More recently, vision-based frameworks such as \cite{li2020super, Super2023,
lu2021superdeepsurgicalperception} have achieved markerless tool tracking by utilizing keypoint detection and a Perspective-n-Point (PnP) solver. \cite{Florian_filter} infused robot kinematic readings with 2D image-based feature detection to jointly estimate surgical instrument pose and kinematic calibration, formulating it as a lumped error. Nevertheless, vision-based feature detection, particularly keypoint detection, remains susceptible to noise and degradation under suboptimal endoscopic imaging conditions, making reliable pose initialization and stable tracking challenging.
Liang \textit{et al.} \cite{liang2025differentiablerenderingbasedposeestimation} first incorporated differentiable rendering into surgical instrument pose reconstruction as a more robust alternative to PnP-based approaches.
Rendering-based frameworks \cite{liang2025differentiablerenderingbasedposeestimation, SurgiPose, yang2025instrumentsplattingcontrollablephotorealisticreconstruction} improve robustness and consistency by fully leveraging differentiable rendering matching in an iterative optimization manner, making them ideal for both single-frame pose reconstruction and kinematic calibration. However, they are computationally slow due to the iterative nature and prone to poor convergence, which limits their applicability in tracking setups with strict time constraints. 

\begin{figure}[t]
\centering
\setlength{\tabcolsep}{1pt}
\renewcommand{\arraystretch}{1.0}

\begin{tabular}{
>{\centering\arraybackslash}m{0.325\columnwidth}
>{\centering\arraybackslash}m{0.325\columnwidth}
>{\centering\arraybackslash}m{0.325\columnwidth}
}

\textbf{Iter 1} &
\textbf{Iter 2} &
\textbf{Iter 3} \\[-1pt]

\includegraphics[width=\linewidth]{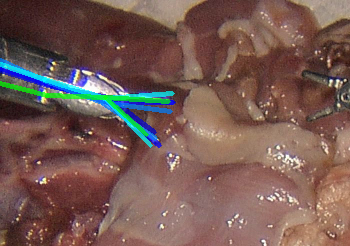} &
\includegraphics[width=\linewidth]{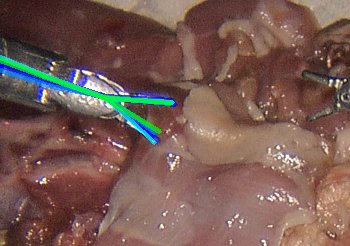} &
\includegraphics[width=\linewidth]{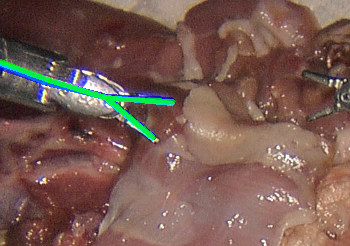} \\

\includegraphics[width=\linewidth]{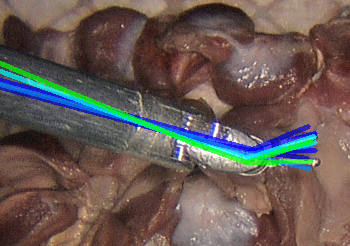} &
\includegraphics[width=\linewidth]{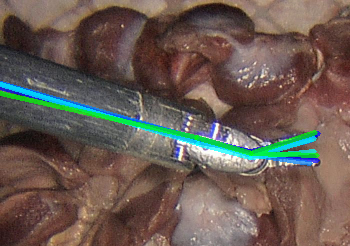} &
\includegraphics[width=\linewidth]{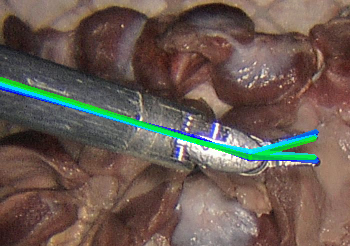}\\

\includegraphics[width=\linewidth]{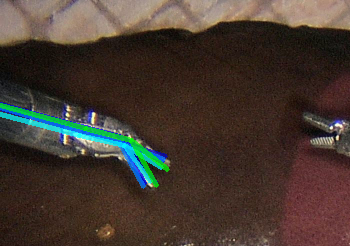} &
\includegraphics[width=\linewidth]{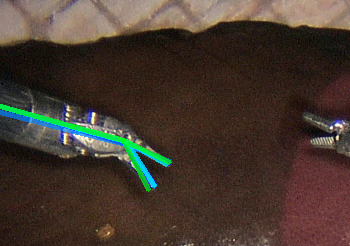} &
\includegraphics[width=\linewidth]{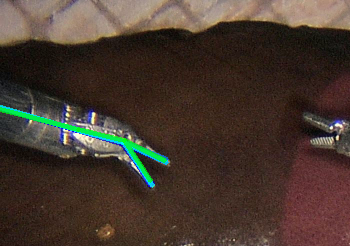}
\end{tabular}
\caption{
Skeleton overlays of the top-$5$ CMA-ES samples across successive iterations. At each iteration, CMA-ES draws a population of candidate poses from a Gaussian distribution, evaluates their fitness using render-and-match objectives, and updates the distribution toward better solutions. Within 3 iterations, the sampled poses concentrate around the correct alignment.
}
\label{fig:sampling_visualization}
\vspace{-0.14in}
\end{figure}

\begin{figure*}[t]
\centering
\includegraphics[width=\textwidth]{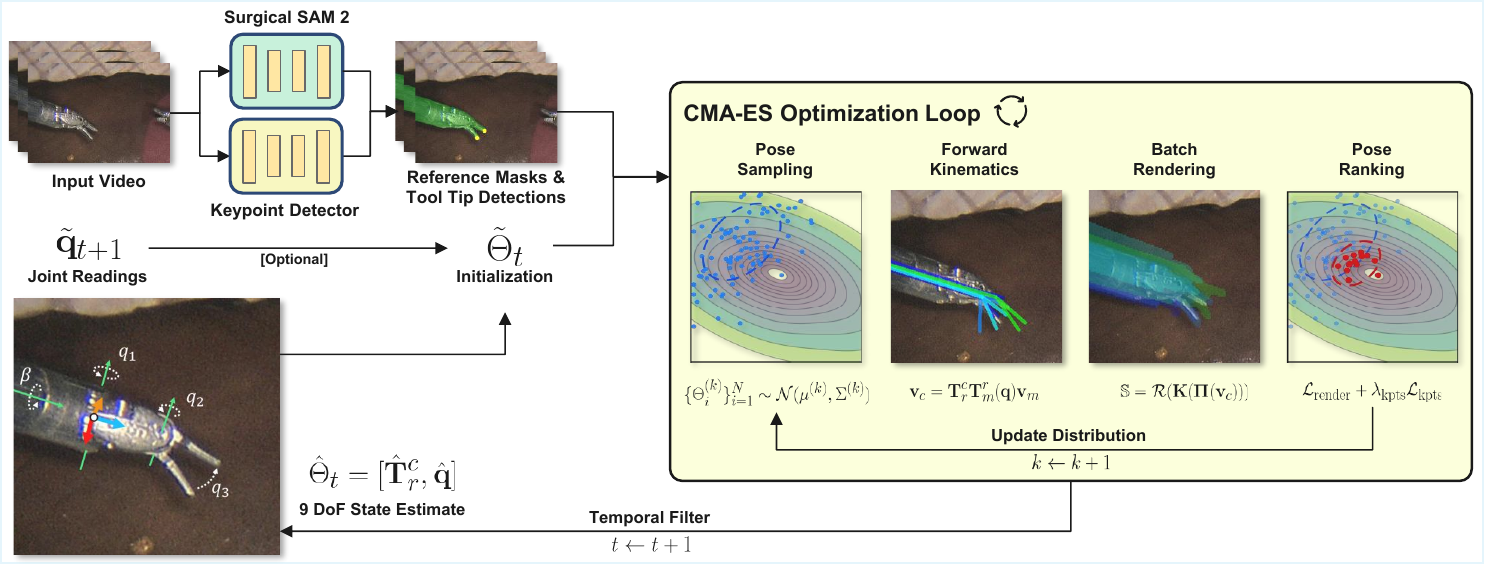}
\caption{\textbf{Overview of the proposed framework.} Given RGB video frames, segmentation masks and tool-tip detections are produced to define a render-and-match objective optimized via CMA-ES. At each iteration, pose candidates are sampled from the current distribution, evaluated in parallel through batched forward kinematics and rendering, and ranked by the objective to update the sampling distribution. The optimized estimates are temporally filtered and propagated to initialize the next frame. In this framework, three joint angles are optimized: wrist pitch $q_1$, wrist yaw $q_2$, and jaw angle $q_3$. The end-effector pose is defined at the wrist pitch frame. A look-at camera representation is adopted to decouple the shaft rotation $\beta$ from the remaining rotational components.}
\label{fig:pipeline}

\vspace{-0.14in}
\end{figure*}

To address these limitations and fully leverage the potential of rendering-based approaches for real-time tracking under noisy kinematic measurements, we propose a novel framework based on evolutionary optimization that jointly estimates surgical instrument pose and joint angles. 
By leveraging the Covariance Matrix Adaptation Evolution Strategy (CMA-ES) \cite{Hansen2001-tj}, which evaluates multiple pose candidates in parallel (Fig.~\ref{fig:sampling_visualization}), the proposed method substantially reduces per-frame iterations and mitigates convergence to local minima, thereby outperforming differentiable rendering. Compared to prior online tracking methods, this framework achieves significant improvements in both accuracy and runtime with GPU-accelerated batch rendering and fitness evaluation.
Furthermore, the framework is extended to handle joint angle-free and bi-manual setups, enabling the calibration of online surgical videos for surgical robot learning applications using only visual inputs. 

The main contributions of this work are as follows:
\begin{enumerate}
    \item A novel integrated online tracking framework that leverages CMA-ES, batch rendering, and temporal filtering to achieve robust real-time surgical instrument pose estimation from a monocular camera.
    \item A tracking formulation that jointly calibrates instrument pose and kinematic measurements, and operates both with and without joint angle inputs, enabling both vision-based feedback control and online surgical video calibration.
    \item An extension to bi-manual tracking that jointly optimizes two articulated instruments within a shared optimization framework, enabling efficient multi-tool tracking with modest computational overhead.
\end{enumerate}

\section{Related Work}


Accurate tracking of articulated surgical robotic instruments remains challenging due to the limited observability of the long kinematic chain, noisy joint readings, and unconventional geometry, typically composed of two small articulated tips and a slender cylindrical shaft extending into the camera view. Early works focused on hardware-based or model-based compensation, including explicit cable-stretch modeling \cite{7353464} and remote center of motion (RCM) calibration \cite{8963603}. There were also approaches including \cite{8460583, 9145695, 9312948, 9484725} that relied on external sensing systems or attached fiducial markers to estimate camera-to-robot transformations and correct kinematic offsets from recorded trajectories. While these methods achieve accurate calibration under controlled conditions, they typically require additional hardware or labor-intensive calibration procedures, making them difficult to deploy and maintain in practical surgical workflows. In addition, physical marker detection is prone to motion blur and self-occlusion. These challenges have driven the recent development of markerless approaches that estimate instrument pose directly from monocular cameras.

Vision-based frameworks for robotic surgical tool tracking typically leverage deep neural networks to detect semantic visual features on the instrument and recover its pose using a PnP solver \cite{8981600, 8932886}. However, for surgical robots such as da Vinci Research Kit (dVRK) \cite{6907809}, pose reconstruction must also account for inaccuracies in joint angle measurements induced by cable-driven transmission. To address this issue, Richter \textit{et al.} \cite{Florian_filter} propose a unified framework that incorporates keypoint detections as observations while estimating a lumped error transformation, enabling simultaneous estimation of the pose and kinematic errors. Building upon this formulation, \cite{10610378} further improves the observation model with deep learning-based cylinder detection. However, these approaches are often unreliable due to the poor endoscopic video quality. They are also limited by the scarcity of data with ground truth annotations. 

More recently, rendering-based approaches have attracted increasing attention due to their consistency and robustness achieved through direct silhouette matching. Liang \textit{et al.} \cite{liang2025differentiablerenderingbasedposeestimation} integrate geometric primitives with differentiable rendering-based optimization to enable robust single-shot calibration. Chen \textit{et al.} \cite{SurgiPose} and Yang \textit{et al.} \cite{yang2025instrumentsplattingcontrollablephotorealisticreconstruction} further apply differentiable rendering to recover surgical instrument trajectories from monocular videos. However, gradient-based optimization in these frameworks remains prone to local minima and slow convergence. Although coarse sampling \cite{liang2025differentiablerenderingbasedposeestimation,  SurgiPose} and image matching \cite{yang2025instrumentsplattingcontrollablephotorealisticreconstruction} strategies have been introduced to improve initialization, the overall optimization process is still computationally expensive. Consequently, existing rendering-based approaches are primarily deployed in offline settings.

\section{Methodology}

An overview of the proposed pipeline is illustrated in Fig.~\ref{fig:pipeline}, which estimates the 6-DoF pose and the visible joint angles of the surgical instrument with respect to the camera frame. Given RGB video frames, instrument segmentation masks are generated in real time using Surgical SAM 2 \cite{liu2024surgicalsam2realtime}, while a learned keypoint detection network localizes the tool tips. These visual observations define a unified optimization objective over pose and joint parameters. The resulting optimization problem is solved using CMA-ES \cite{Hansen2001-tj}, an evolution strategy that iteratively refines a Gaussian search distribution over candidate states, with efficient objective evaluations carried out through batch rendering. A Kalman filter with a constant-velocity motion model is incorporated to enforce temporal consistency, and the filtered state is used to initialize the search distribution for the subsequent frame. 

\subsection{Surgical Instrument Tracking}

This study considers the Patient Side Manipulator (PSM) equipped with a Large Needle Driver (LND) from the da Vinci\textregistered\ Surgical System. The manipulator has 6 DoF, along with an additional gripper angle. As shown in Fig.~\ref{fig:pipeline}, the end-effector coordinate frame is defined at the wrist pitch frame (joint 4) of the PSM. Similar to prior works \cite{DR_orig, liang2025differentiablerenderingbasedposeestimation}, the surgical instrument localization is formulated as a nonlinear optimization problem over the end-effector pose $\mathbf{T}_{r}^c \in SE(3)$, together with the last three visible joint angles $\mathbf{q} = [q_1, q_2, q_3]^\top \in \mathbb{R}^3$ corresponding to wrist pitch, wrist yaw, and gripper angle. For each RGB video frame $\mathbb{I} \in \mathbb{R}^{H \times W \times 3}$, the objective
\begin{equation}
    \Theta^\ast = \argmin_\Theta \mathcal{L}(\mathbb{I}, \Theta)
    \label{eqn:objective}
\end{equation}
is to estimate the state $\Theta = (\mathbf{T}_{r}^c, \mathbf{q})$ that minimizes the discrepancy between observed features from $\mathbb{I}$ (e.g., segmentation masks and keypoints) and the reconstructions produced from the estimated state $\Theta$. After a single-shot calibration to initialize the solution, the state estimate from the previous frame is propagated to initialize optimization at the current frame. When joint angle readings are available, the measured values are used to initialize the optimization rather than relying on the propagated estimation. 

\subsubsection{Evolutionary Optimization}

Rather than relying on gradient-based optimization, the Covariance Matrix Adaptation Evolution Strategy (CMA-ES) \cite{Hansen2001-tj} is adopted. CMA-ES is a blackbox optimization method that iteratively refines a search distribution over the state space by maximizing the expected fitness $J(\mu, \Sigma) := \mathbb{E}\big[-\mathcal{L}(\mathbb{I}, \Theta)\mid \mu, \Sigma\big]$, leading to the optimal parameters
\begin{equation}
(\mu^\ast, \Sigma^\ast) = \argmax_{\mu, \Sigma} J(\mu, \Sigma)
\label{eqn:expected_fitness}
\end{equation}
where $\mu, \Sigma$ defines a Gaussian distribution over the state variable $\Theta$. This formulation effectively lifts the original objective $\mathcal{L}(\mathbf{I}, \Theta)$ from the parameter level to the distribution level, such that the loss landscape with respect to the mean $\mu$ approximates a Gaussian-smoothed version of $\mathcal{L}(\mathbf{I}, \Theta)$, improving robustness to local minima; and covariance adaptations enable rapid convergence to locally quadratic optima of the expected fitness $J(\mu, \Sigma)$ \cite{XNES}.

A key advantage of CMA-ES in rendering-based pose estimation is its natural compatibility with blackbox batch rendering. At each iteration, the search distribution is updated based on the top-$M$ elite candidates $\{\Theta_i^{(k)}\}_{i=1}^M$ of the $N$ samples generated from the previous distribution $\mathcal{N}(\mu^{(k)}, \Sigma^{(k)})$
\begin{align}
    \mu^{(k+1)} &= \mu^{(k)} + c_m \sum_{i=1}^{K} w_i (\Theta_i^{(k)} - \mu^{(k)}), \\
\begin{split}
    \Sigma^{(k+1)} &= (1-c_1-c_\mu)\Sigma^{(k)} + c_1 \mathbf{p}_c \mathbf{p}_c^\top \\
    &\qquad + c_\mu \sum_{i=1}^{K} w_i \frac{\Theta_i^{(k)} - \mu^{(k)}}{\sigma_k}\left(\frac{\Theta_i^{(k)} - \mu^{(k)}}{\sigma_k}\right)^\top
\end{split}
\label{eqn:cmaes_update}
\end{align}
where $\{w_i\}_{i=1}^K$ are the weights of the elite candidates, $c_m$, $c_1$, and $c_\mu$ are learning rates, $\mathbf{p}_c$ is the evolution path that accumulates successive steps of $\mu$ to introduce momentum, and $\sigma_k$ is the adapted step size. These updates depend only on the fitness ranking of sampled states, eliminating the need for differentiable rendering pipelines and effectively exploiting batch rendering to evaluate multiple candidates in parallel at each iteration, substantially reducing the wall-clock time for pose estimation compared to gradient-based methods.

\subsubsection{Loss Function}

The objective function in (\ref{eqn:objective}) combines rendering loss as well as robust keypoint alignment loss to provide additional geometric constraints
\begin{equation}
    \mathcal{L}\big(\mathbb{I}, \Theta\big) = \mathcal{L}_\text{render} + \lambda_\text{kpts} \mathcal{L}_\text{kpts}
\end{equation}
where the rendering loss includes the Mean Squared Error (MSE) loss and the appearance loss \cite{DR_orig, liang2025differentiablerenderingbasedposeestimation} to adjust the distance of the robot model
\begin{align}
    \begin{split}
        \mathcal{L}_\text{render}(\mathbb{M}_\text{ref}, \mathbb{S}) &= \underbrace{\sum_{i=0}^{H-1} \sum_{j=0}^{W-1} (\mathbb{S}(i,j)-\mathbb{M}_\text{ref}(i,j))^2}_{\text{MSE loss}} \\
        &\quad+ \lambda_\text{app}\underbrace{\left\|\sum_{i=0}^{H-1} \sum_{j=0}^{W-1}\mathbb{S}(i,j)-\mathbb{M}_\text{ref}(i,j)\right\|}_{\text{Appearance Loss}}.
    \end{split}
    \label{eqn:render_loss}
\end{align}
The segmentation $\mathbb{M}_{\text{ref}}$ is produced by Surgical SAM 2 \cite{liu2024surgicalsam2realtime} given the RGB image $\mathbb{I}$. The rendered mask $\mathbb{S} \in \{0, 1\}^{H \times W}$ of the surgical tool is obtained from a renderer $\mathcal{R}(\cdot)$ given the mesh vertices transformed to the camera frame
\begin{equation}
    \mathbb{S} = \mathcal{R}(\mathbf{K}(\Pi(\mathbf{T}_{r}^c \mathbf{T}_m^r(\mathbf{q})\mathbf{v}_m)))
\end{equation}
where $\mathbf{K} \in \mathbb{R}^{3\times 3}$ is the camera intrinsic matrix, $\Pi(\cdot)$ is the canonical projection, $\mathbf{T}_m^r(\mathbf{q})$ is the forward kinematics transform from mesh frame to robot frame, and $\mathbf{v}_m \in \mathbb{R}^{n\times 3}$ denotes the mesh vertices in the CAD mesh frame.

A heatmap-based 2D keypoint detection network with a ResNet-18 backbone is trained to localize the tool tips from the mask $\mathbb{S}$. Denoting the detected tool tips as $\mathbf{t}_1^r, \mathbf{t}_2^r$ and the corresponding projected keypoints as $\mathbf{t}_1^p, \mathbf{t}_2^p$ obtained from the pose candidates, the keypoint loss $\mathcal{L}_\text{kpts}$ is given by
\begin{align}
\begin{split}
    \mathcal{L}_{\text{kpts}}(\mathbf{t}^r,\mathbf{t}^p) &=\min_{\sigma \in \{\{1,2\},\,\{2,1\}\}}\sum_{i=1}^{2}\left(\left\|\mathbf{t}_i^{r}-\mathbf{t}_{\sigma(i)}^{p}\right\|_2 -\tau\right)_+  \\
    &\quad + (\left\|\mathbf{t}_{\text{mean}}^{r}-\mathbf{t}_{\text{mean}}^{p}\right\|_2-\tau)_+
    \label{eqn:kpts_loss}
\end{split}
\end{align}
where $(x)_+ = \max(0,x)$ is the positive-part operator and $\tau > 0$ is a threshold that mitigates the effect of keypoint localization noise. The keypoint loss $\mathcal{L}_\text{kpts}$ is evaluated only when exactly two distinct keypoints are decoded from the heatmap, ensuring robustness to corrupted segmentation masks that may cause missed tips or spurious detections.

\subsubsection{State Parameterization}

A unified rotation representation introduced by Liang \textit{et al.} \cite{liang2025differentiablerenderingbasedposeestimation} is adopted for pose sampling within CMA-ES. Specifically, the look-at camera representation $[\alpha, \beta, \gamma]^\top \in \mathbb{R}^3$ is used, such that
\begin{align}
\mathbf{R}_{\text{LookAt}}(\alpha, \beta, \gamma)
= \mathbf{R}_y(\gamma)\mathbf{R}_x(\alpha)\mathbf{R}_z(\beta),
\end{align}
where $\mathbf{R}_x(\cdot)$, $\mathbf{R}_y(\cdot)$, and $\mathbf{R}_z(\cdot)$ are rotations about the $x$-, $y$-, and $z$-axes, respectively. The pose transformation from the robot frame to the camera frame is defined as
\begin{align}
\mathbf{T}_r^c(\alpha, \beta, \gamma, x, y, z) =
\begin{bmatrix}
\mathbf{R}_{\text{LookAt}}(\alpha, \beta, \gamma) & \mathbf{t} \\
\mathbf{0}^\top & 1
\end{bmatrix},
\end{align}
where $\mathbf{t} = [x, y, z]^\top \in \mathbb{R}^3$ is the translation vector. This formulation explicitly decouples the shaft rotation $\beta$ from the remaining rotational components (Fig.~\ref{fig:pipeline}), enabling structured pose sampling and accommodating the $180^\circ$ rotational symmetry of instruments such as the Large Needle Driver.

In addition, joint angles are subject to physical box constraints, while CMA-ES does not natively support constrained optimization. Hard clamping or sigmoid-based mappings, although commonly used in gradient-based methods, can distort the sampling distribution and introduce bias toward boundary values when applied within the evolution strategy \cite{caraffini2019infeasibility}. A cosine-based reparameterization is adopted to enforce the constraints while maintaining a well-behaved search distribution. Denoting the lower and upper bounds of the $i$-th joint angle as $q^\text{lb}_{i}$ and $q^\text{ub}_{i}$, the mapping from a valid joint angle reading $q_i \in [q^\text{lb}_{i}, q^\text{ub}_{i}]$ to an unconstrained representation $\hat{q}_i$ for CMA-ES optimization is given by 
\begin{equation}
    \hat{q}_i = q^\text{lb}_{i} + \frac{q^\text{ub}_{i}-q^\text{lb}_{i}}{\pi} \arccos\left( 1 - 2 \frac{q_i - q^\text{lb}_{i}}{q^\text{ub}_{i}-q^\text{lb}_{i}} \right),
\end{equation}
and the inverse mapping that recovers a valid joint angle $q_i$  for fitness evaluation from the unconstrained variable $\hat{q}_i$ is
\begin{equation}
    q_i = q^\text{lb}_{i} + \frac{q^\text{ub}_{i}-q^\text{lb}_{i}}{2} \left( 1 - \cos\left(\pi \, \frac{\hat{q}_i - q^\text{lb}_{i}}{q^\text{ub}_{i}-q^\text{lb}_{i}}\right) \right).
\end{equation}
This reparameterization guarantees that all CMA-ES samples respect the joint limits while maintaining an unconstrained search distribution internally.

\subsubsection{Temporal Filtering}

A temporal filter is applied to the state estimates to reduce optimization-induced jitter and mitigate noise from segmentation masks and keypoint detection. Specifically, the filter is defined as
$\mathbf{s}_t = [\Theta_t, \dot{\Theta}_t]^\top \in \mathbb{R}^{18}$, 
where 
the 9-DoF state $\Theta_t$
encodes the pose and joint angles, and $\dot{\Theta}_t\in \mathbb{R}^9$ their corresponding velocities.  
The motion model assumes constant velocity
\begin{equation}
\mathbf{s}_t = 
\begin{bmatrix} \mathbf{I} & \Delta t \mathbf{I} \\ \mathbf{0} & \mathbf{I} \end{bmatrix}
\mathbf{s}_{t-1} + \mathbf{w}_t, \quad 
\mathbf{w}_t \sim \mathcal{N}(\mathbf{0}, \mathbf{Q}),
\label{eqn:kf_motion}
\end{equation}
where $\mathbf{I}, \mathbf{0}$ denote the identity and zero matrices, $\Delta t$ is the time interval between frames, $\mathbf{w}_t$ is the process noise, and $\mathbf{Q} \in \mathbb{R}^{18 \times 18}$ is the process noise covariance. Furthermore, estimates from CMA-ES optimization serve as direct observations $\mathbf{z}_t$ of the state
\begin{equation}
\mathbf{z}_t = 
[\mathbf{I} \;\; \mathbf{0}]
\mathbf{s}_t + \mathbf{v}_t, \quad 
\mathbf{v}_t \sim \mathcal{N}(\mathbf{0}, \mathbf{R}),
\end{equation}
where $\mathbf{v}_t$ is the observation noise and $\mathbf{R} \in \mathbb{R}^{9\times 9}$ is the measurement noise covariance. Filtered joint angles are clamped to their physical limits to ensure valid configurations.

\subsection{Bi-Manual Tracking}

\begin{figure*}[t]
\centering
\setlength{\tabcolsep}{1pt}
\renewcommand{\arraystretch}{1.05}

\begin{tabular}{
>{\centering\arraybackslash}m{0.095\textwidth}
>{\centering\arraybackslash}m{0.175\textwidth}
>{\centering\arraybackslash}m{0.175\textwidth}
>{\centering\arraybackslash}m{0.175\textwidth}
>{\centering\arraybackslash}m{0.175\textwidth}
>{\centering\arraybackslash}m{0.175\textwidth}
}

&
\textbf{$t = 250$} &
\textbf{$t = 300$} &
\textbf{$t = 350$} &
\textbf{$t = 400$} &
\textbf{$t = 450$}\\[0pt]

\shortstack{\textbf{GD}  \\ \footnotesize w/o joint \\ \footnotesize 10 iter/frame} &
\includegraphics[width=\linewidth]{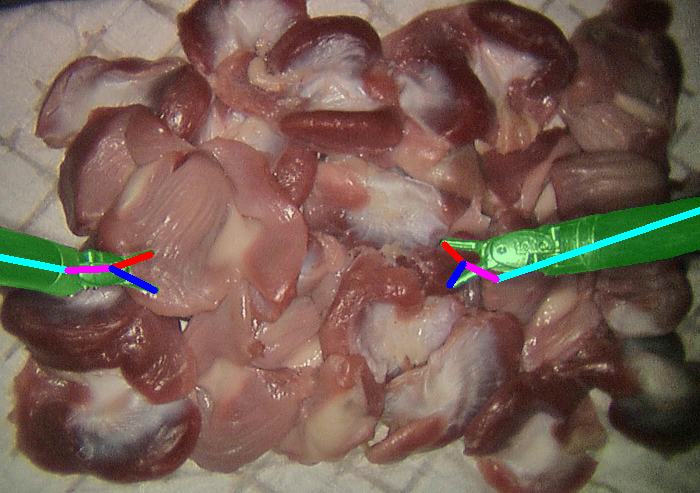}&
\includegraphics[width=\linewidth]{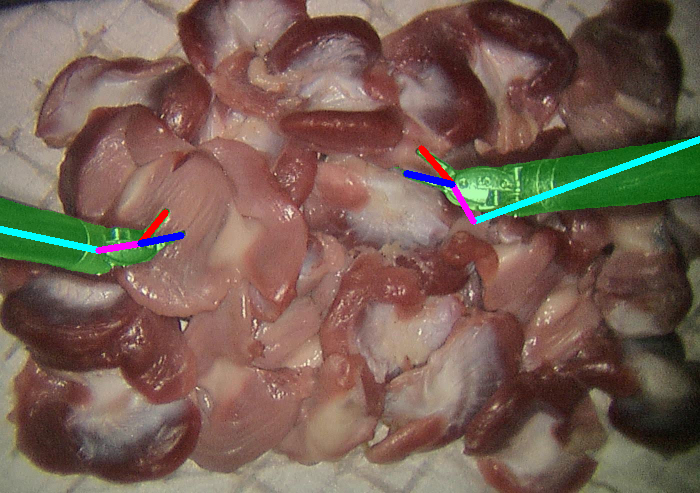} &
\includegraphics[width=\linewidth]{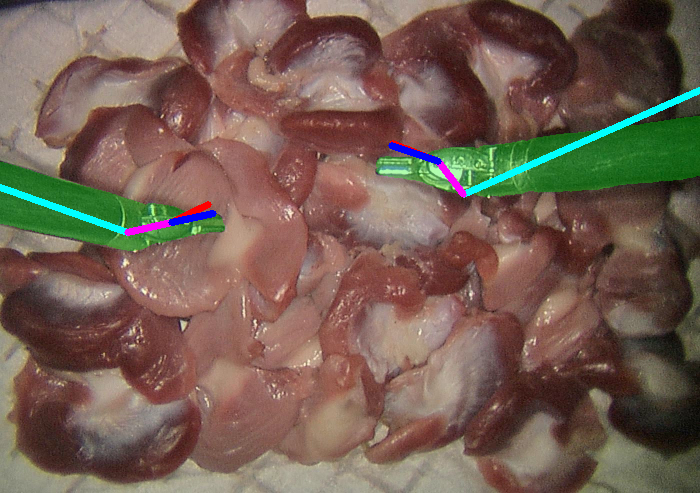} &
\includegraphics[width=\linewidth]{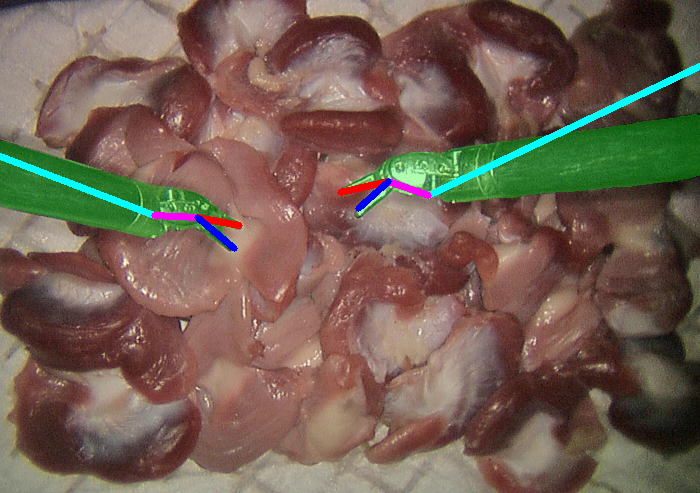} & 
\includegraphics[width=\linewidth]{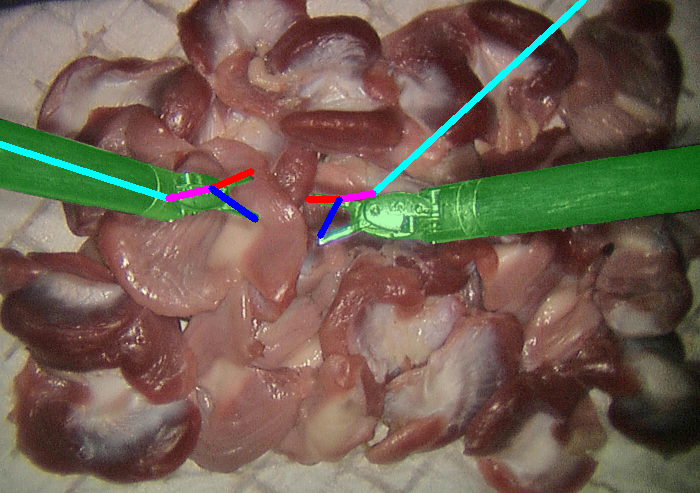} \\[8pt]

\shortstack{\textbf{Ours}  \\ \footnotesize w/o joint \\ \footnotesize 3 iter/frame} &
\includegraphics[width=\linewidth]{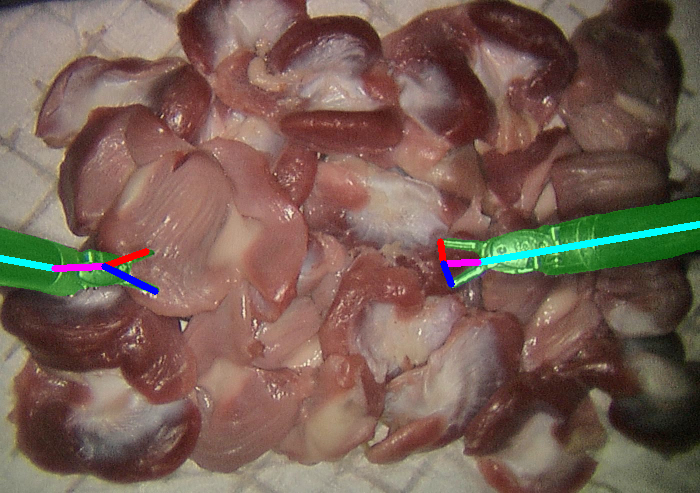} &
\includegraphics[width=\linewidth]{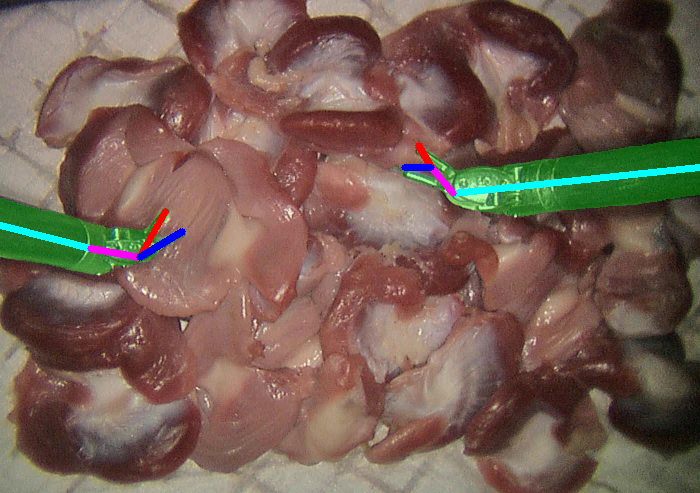} &
\includegraphics[width=\linewidth]{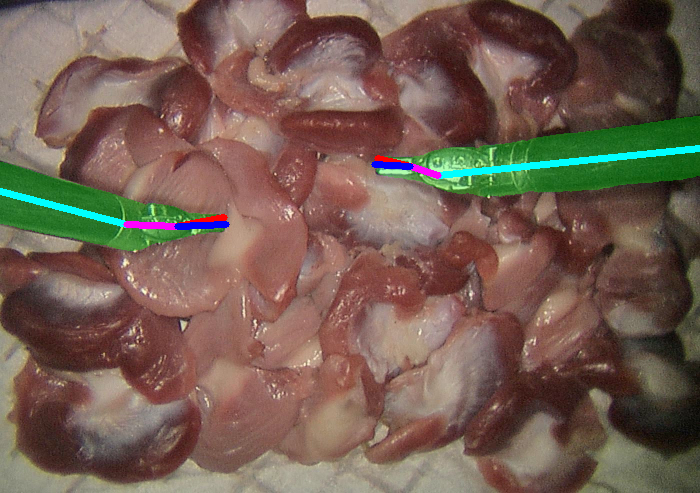} &
\includegraphics[width=\linewidth]{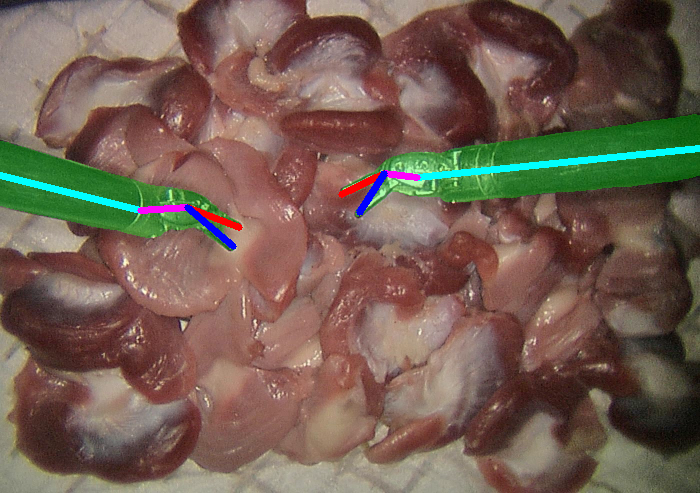} & 
\includegraphics[width=\linewidth]{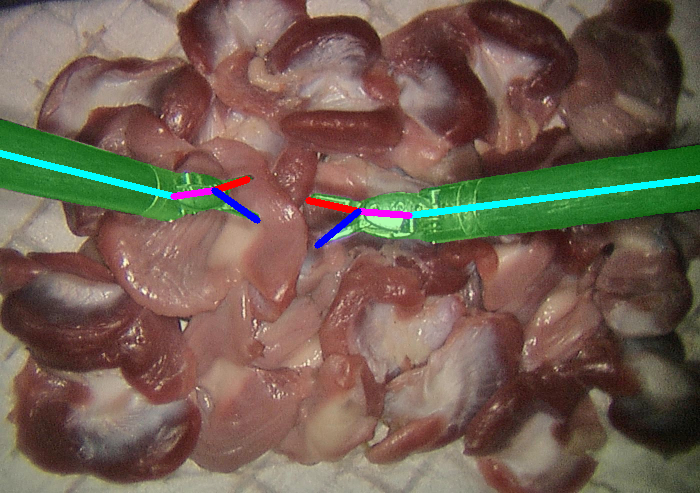}\\[8pt]

\shortstack{\textbf{Ours} \\ \footnotesize w/ joint \\ \footnotesize 3 iter/frame} &
\includegraphics[width=\linewidth]{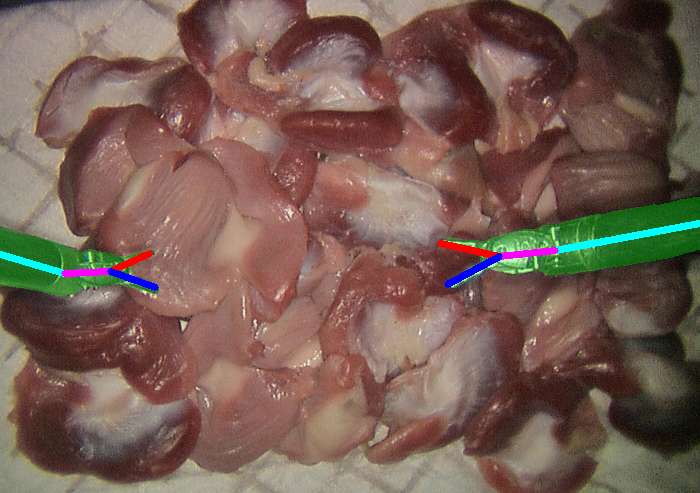} & 
\includegraphics[width=\linewidth]{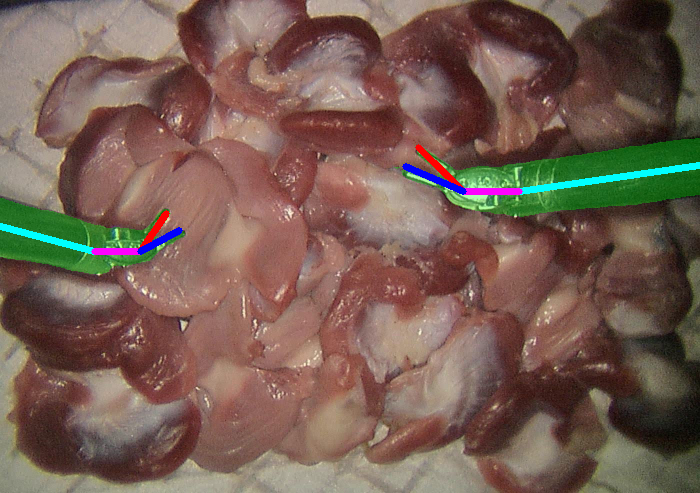} &
\includegraphics[width=\linewidth]{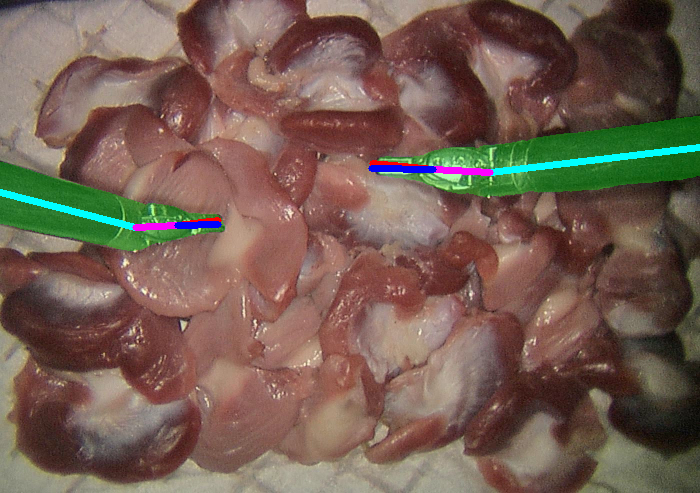} &
\includegraphics[width=\linewidth]{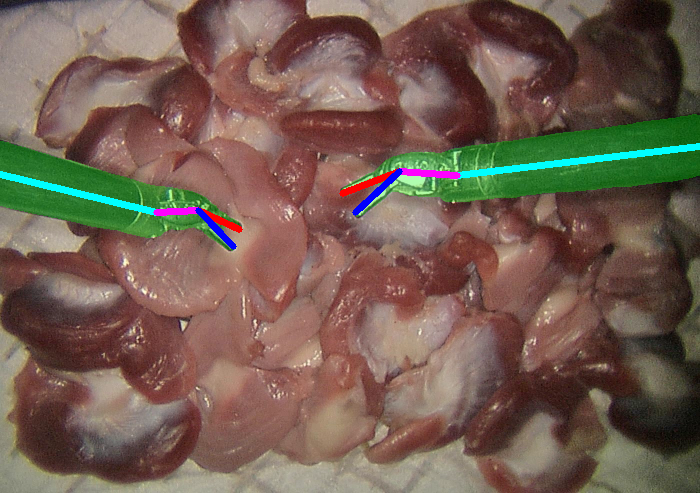} &
\includegraphics[width=\linewidth]{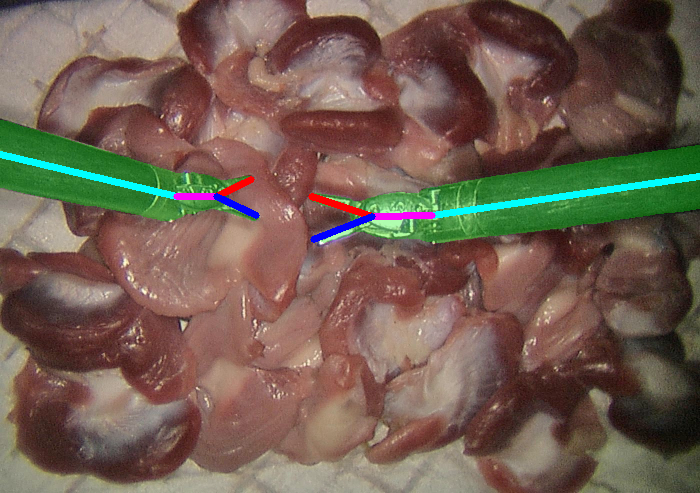}

\end{tabular}

\vspace{-0.06in}

\caption{Qualitative comparison of bi-manual tracking on the SurgPose dataset. Reference masks are shown with green overlays. The proposed method achieves accurate pose reconstruction when joint angle readings are available and remains robust to poor initialization even without joint measurements, while the gradient-based approach is prone to local minima and error accumulation.}

\label{fig:qualitative_plot_1}
\vspace{-0.09in}
\end{figure*}

\begin{figure*}[t]
\centering
\setlength{\tabcolsep}{1pt}
\renewcommand{\arraystretch}{1.05}

\begin{tabular}{
>{\centering\arraybackslash}m{0.095\textwidth}
>{\centering\arraybackslash}m{0.175\textwidth}
>{\centering\arraybackslash}m{0.175\textwidth}
>{\centering\arraybackslash}m{0.175\textwidth}
>{\centering\arraybackslash}m{0.175\textwidth}
>{\centering\arraybackslash}m{0.175\textwidth}
}

&
\textbf{$t = 100$} &
\textbf{$t = 110$} &
\textbf{$t = 120$} &
\textbf{$t = 130$} &
\textbf{$t = 140$}\\[0pt]

\shortstack{\textbf{Particle} \\ \textbf{Filter} \\ \footnotesize 1k particles} &
\includegraphics[width=\linewidth]{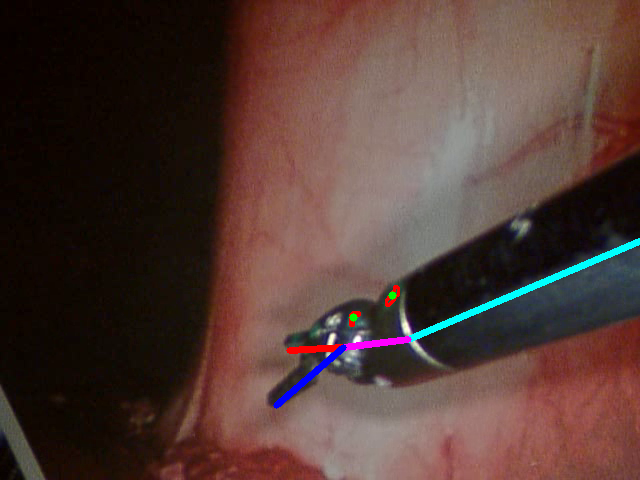}&
\includegraphics[width=\linewidth]{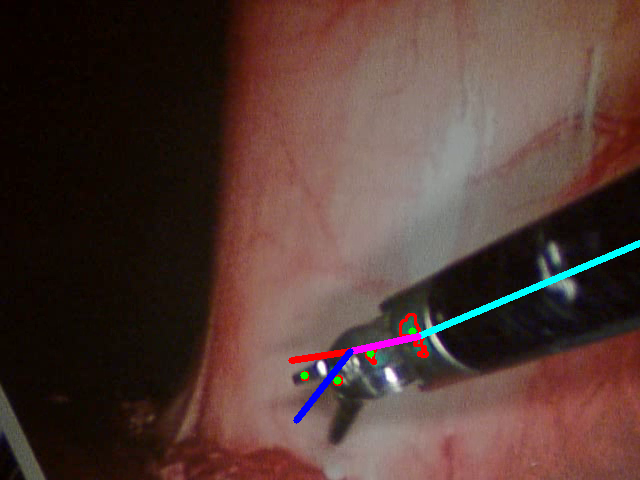} &
\includegraphics[width=\linewidth]{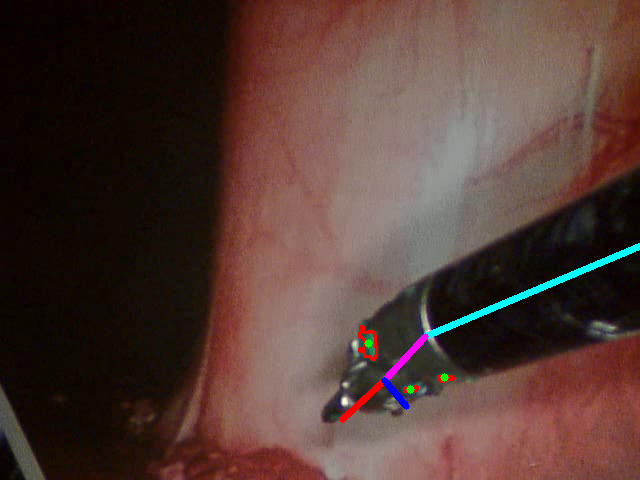} &
\includegraphics[width=\linewidth]{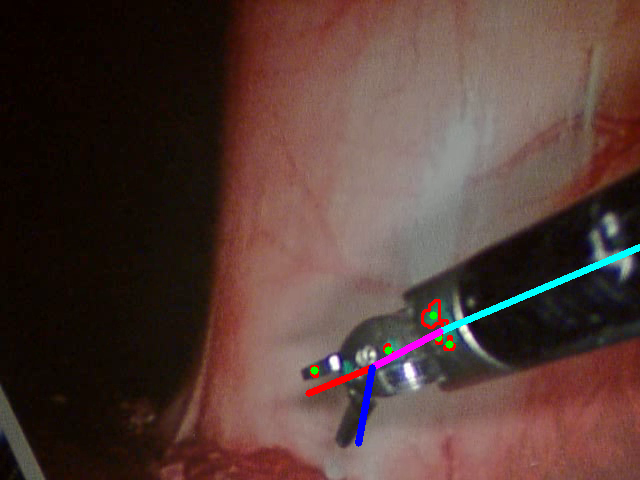} & 
\includegraphics[width=\linewidth]{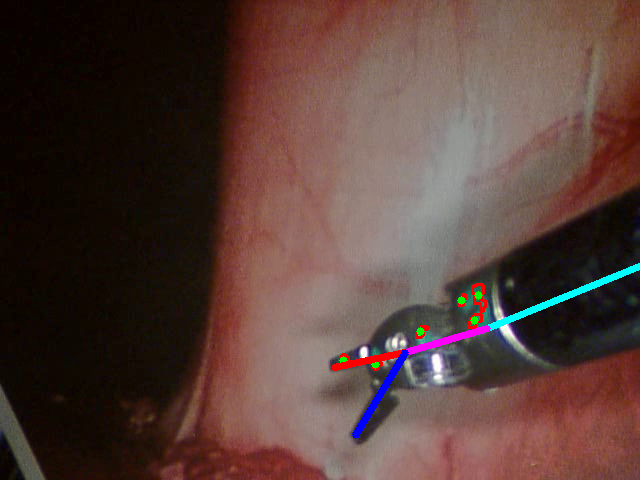} \\[8pt]

\shortstack{\textbf{Ours}  \\ \footnotesize w/ joint \\ \footnotesize 3 iter/frame} &
\includegraphics[width=\linewidth]{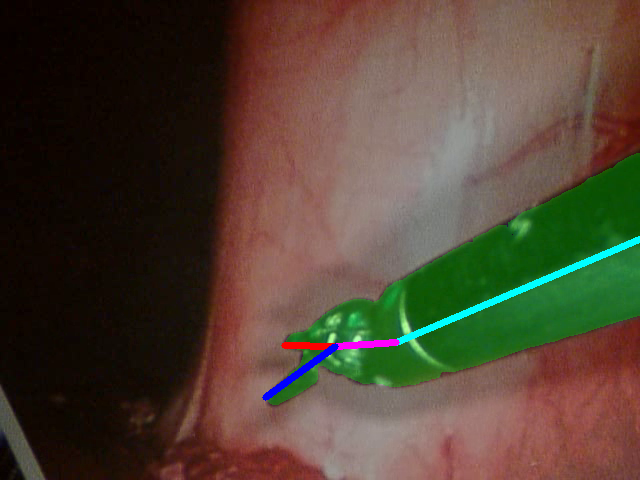} &
\includegraphics[width=\linewidth]{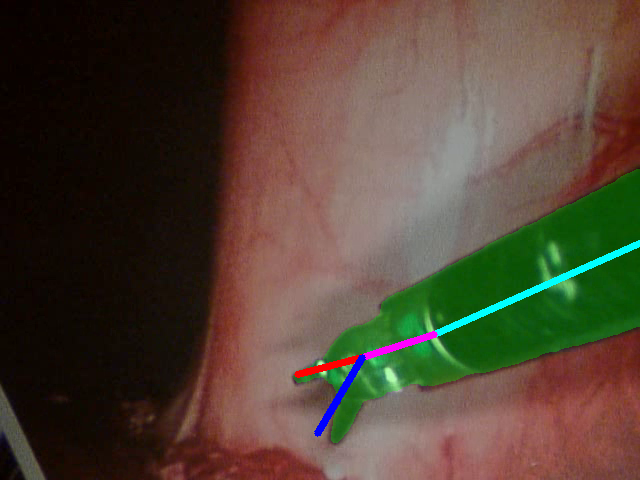} &
\includegraphics[width=\linewidth]{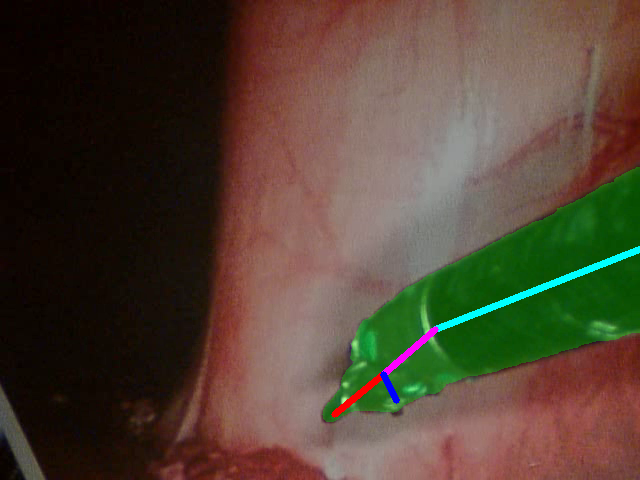} &
\includegraphics[width=\linewidth]{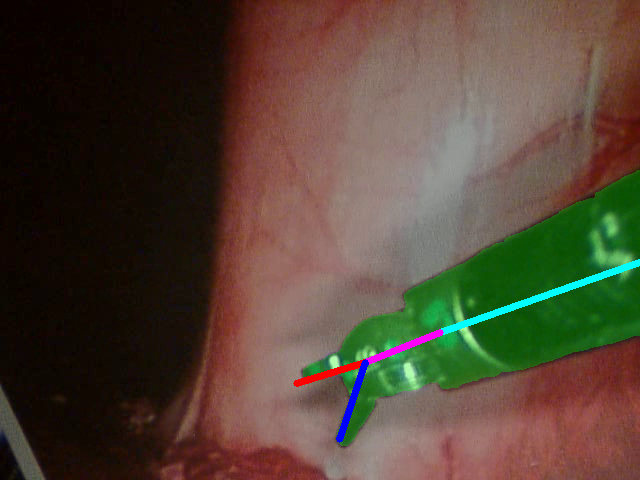} & 
\includegraphics[width=\linewidth]{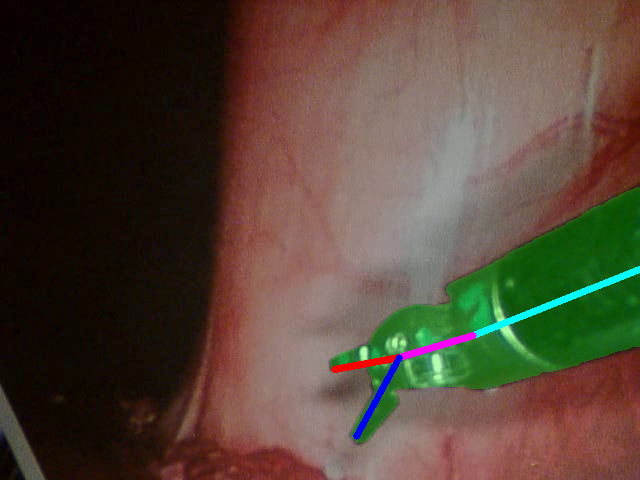}\\

\end{tabular}

\vspace{-0.06in}

\caption{Qualitative comparison of single-arm tracking on the collected dataset. Detected colored markers for the particle filter are shown as green dots (centroids of the regions outlined by red contours). Reference masks for the proposed approach are shown with green overlays. Our method demonstrates improved accuracy in the alignment of tool tips compared to the particle filter.
}

\label{fig:qualitative_plot_2}
\vspace{-0.14in}
\end{figure*}


To extend the proposed framework to bi-manual tracking setups, the state parameters for both arms are jointly optimized as a concatenated parameter $\Theta = [\Theta_L, \Theta_R]^\top \in \mathbb{R}^{18}$, where $\Theta_L, \Theta_R \in \mathbb{R}^9$ represent the 9-DoF state of the left and right arms. In this formulation, both arms are rendered jointly, and the rendering loss (\ref{eqn:render_loss}) is evaluated on the combined image. The overall objective is then defined as:
\begin{align}
    \mathcal{L}(\mathbb{I}, \Theta) = \mathcal{L}_\text{render} + \lambda_\text{kpts} (\mathcal{L}_\text{kpts}^L + \mathcal{L}_\text{kpts}^R),
    \label{eqn:bi_manual_loss}
\end{align}
where $\mathcal{L}_\text{kpts}^L$ and $\mathcal{L}_\text{kpts}^R$ denote the keypoint losses for the left and right arms, respectively.



Inspired by separable CMA-ES \cite{sepCMAES}, we constrain the covariance update (\ref{eqn:cmaes_update}) in CMA-ES to a block-diagonal form:
\begin{equation}
\Sigma^{(k)} =
\begin{bmatrix}
\Sigma_L^{(k)} & 0 \\
0 & \Sigma_R^{(k)}
\end{bmatrix},
\end{equation}
so that the evolution of each arm's state is governed independently by its own covariance block $\Sigma_L^{(k)}, \Sigma_R^{(k)} \in \mathbb{R}^{9\times 9}$. This design reduces model complexity and prevents spurious correlations between the states of the two arms. 

\section{Experiments}

\subsection*{Implementation Setups}

All experiments are conducted on a workstation equipped with an NVIDIA RTX 4090 GPU and an AMD Ryzen 9 9960X CPU. Reference segmentation masks used in the tracking pipeline are generated by Surgical SAM 2 \cite{liu2024surgicalsam2realtime} with point prompts provided in the first frame of the trajectory. The optimization frameworks are implemented with EvoTorch \cite{evotorch} and NvDiffRast \cite{Laine2020diffrast}. For evolutionary optimization, gradients are not evaluated, and instance-mode batch rendering is used for GPU-parallelized evaluation of the generated samples in each iteration. 

For initialization, the pose-ranking method of Liang \textit{et al.} \cite{liang2025differentiablerenderingbasedposeestimation} is adapted to enable a fully joint angle-free pipeline. In addition to the original pose hypothesis space, initial joint configurations are also sampled from a feasible set. Instead of uniformly sampling this expanded search space, TuRBO \cite{eriksson2019scalable} is applied to adaptively concentrate exploration on promising regions, thereby improving efficiency despite the increased dimensionality introduced by the additional joint variables. 

During tracking, the parameter space is normalized to account for heterogeneous units and magnitudes. CMA-ES is initialized with covariance $\Sigma^{(0)} = \mathbf{I}$, and the default learning rates $c_m$ and $c_\mu$ are increased to accelerate convergence, with $N = 70$ samples generated per iteration. The Adam optimizer \cite{kingma2017adammethodstochasticoptimization} with a learning rate of $0.1$ in the normalized space serves as a gradient-based baseline.

\subsection*{Synthetic Dataset}

To quantitatively evaluate the pose reconstruction accuracy, 16 pairs of synthetic trajectories (left and right arms) are generated. Each pair consists of 1000 frames and is constructed by selecting 20 feasible waypoints and interpolating between them using cubic Hermite splines. The trajectories are then perturbed with temporally correlated noise via an Ornstein-Uhlenbeck process and smoothed with a low-pass filter. Corresponding binary masks are rendered at a resolution of $700 \times 493$. To simulate realistic segmentation errors, the masks are further perturbed with random dilation, erosion, and edge blobs. Additionally, 10,000 synthetic masks are generated from the pose hypothesis space to train the keypoint detector.

\subsection*{Real-world Dataset}

\begin{figure*}[t]
    \centering
    \includegraphics[width=0.93\textwidth]{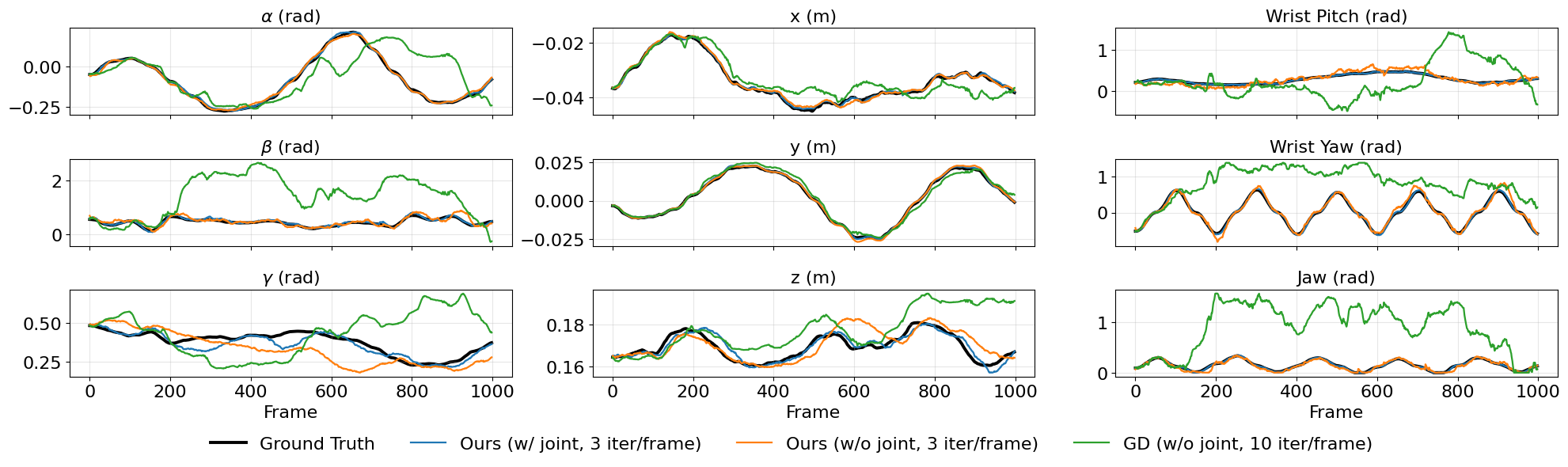}
    \caption{Qualitative comparison of pose reconstruction on a synthetic trajectory. The proposed methods produce more stable and accurate estimates than the gradient-based baseline. The larger errors in the $\gamma$ and $z$ components are likely due to weak silhouette constraints: $\gamma$ is mainly inferred from subtle shaft-boundary changes, while depth estimation is sensitive to mask-size errors.}
    \label{fig:synthetic_qual}
\vspace{-0.14in}
\end{figure*}

For real-world evaluation, 12 trajectories from the SurgPose dataset \cite{wu2025surgposedatasetarticulatedrobotic} are selected, with both manipulators equipped with Large Needle Drivers (LND). The dataset is divided into 8 sequences with nominal speed and 4 sequences exhibiting rapid movements. In addition, we collected 6 trajectories of synchronized left and right stereo image streams at $640 \times 480$ with painted markers and joint angle readings to benchmark against the online tool tracking method proposed by Richter \textit{et al.} \cite{Florian_filter}, which relies on colored markers on the instrument for keypoint detection and applies a particle filter to estimate the lumped error transform. The collected dataset also includes refined segmentation masks to enable more accurate quantitative evaluation. The recorded trajectories vary in length from 250 to 554 frames, with an average of 436 frames per sequence. For the fast-motion trajectories in SurgPose and in the collected dataset, the process noises of the velocity term in the Kalman filter motion model~(\ref{eqn:kf_motion}) are correspondingly increased to better accommodate larger inter-frame velocity deviations.

\subsection{Qualitative Results.} 

The qualitative tracking results are shown in Fig.~\ref{fig:qualitative_plot_1} and Fig.~\ref{fig:qualitative_plot_2}, illustrating the accuracy and robustness of different tracking strategies on both the SurgPose dataset and the collected dataset. On the SurgPose dataset (Fig.~\ref{fig:qualitative_plot_1}), the proposed method is compared against the gradient-based baseline for bi-manual tracking. Gradient descent suffers from stagnation in local optima, which leads to error accumulation and visible drift over time. In contrast, CMA-ES demonstrates stronger robustness to early tracking errors (e.g., at Frame $250$) even without joint angle readings. Furthermore, the tracking becomes more accurate when joint angle readings are incorporated, producing consistent alignment between the estimated pose and image observations across frames. Figure~\ref{fig:qualitative_plot_2} presents the comparison with the particle filter baseline of Richter \textit{et al.} \cite{Florian_filter} on the collected dataset. Using 1000 particles as in their original configuration, the particle filter shows noticeable misalignment at the tool tips. In contrast, the proposed method achieves more accurate alignment while maintaining smoother and more stable tracking.

On the synthetic dataset (Fig.~\ref{fig:synthetic_qual}), the estimated rotation, translation, and joint trajectories are compared against ground truth. CMA-ES with joint angle readings achieves the most accurate reconstruction across all components, closely matching the reference motion throughout the sequence. Even without joint measurements, the proposed method maintains stable and well-aligned estimates, consistently outperforming the gradient-based baseline. In contrast, gradient descent exhibits noticeable deviations, particularly in rotation and joint angle trajectories, reflecting its sensitivity to local minima and reduced robustness over time.

\subsection{Quantitative Results.} 
\newcommand{\tabspacerow}{  
    \addlinespace[-4pt]
    \multicolumn{13}{l}{} \\
    \addlinespace[-4pt]
}

\newcommand{\tablesection}[1]{%
  \hline
  \addlinespace[2pt]
  \multicolumn{13}{l}{\textbf{#1}} \\
  \addlinespace[2pt]
  \hline
}

\newcommand{\xmark}{} 
\newcommand{\bcheckmark}{\scalebox{1.0}{\checkmark}}
\newcommand{\bxmark}{\scalebox{1.0}{\xmark}}

\setlength{\tabcolsep}{3pt}
\renewcommand{\arraystretch}{1.05}

\begin{table*}[t]
\centering
\resizebox{\textwidth}{!}{
\begin{tabular}{c
| c c
| c
| c c c c c
| c c
| c c}

\hline

\tabspacerow

\textbf{Optimizer}
& \multicolumn{2}{c|}{\textbf{Setup}}
& \multicolumn{1}{c|}{\textbf{Speed}}
& \multicolumn{5}{c|}{\textbf{Synthetic Data}}
& \multicolumn{4}{c}{\textbf{SurgPose}} \\


& \multirow{2}{*}{\centering Joint}
& \multirow{2}{*}{\centering Iters}
& \multirow{2}{*}{\centering Time (ms)}
& \multirow{2}{*}{\centering Rot.}
& \multirow{2}{*}{\centering Trans.}
& \multirow{2}{*}{\centering $q_1$}
& \multirow{2}{*}{\centering $q_2$}
& \multirow{2}{*}{\centering $q_3$}
& \multicolumn{2}{c|}{Normal}
& \multicolumn{2}{c}{Fast} \\

& & & & & & & &
& Mask Err.
& KP Err.
& Mask Err.
& KP Err. \\

\tabspacerow


\tablesection{Single-Arm Tracking}
\tabspacerow

\multirow{3}{*}{\centering GD (NvDiffRast)}
& \xmark & 10
& 42.40
& 0.4957 & 0.0089 & 0.3378 & 0.3377 & 0.1715
& 0.1432 & 8.8057
& 0.2650 & 8.5244 \\

& \bcheckmark & 10
& 42.51
& 0.2383 & 0.0052 & \underline{0.0326} & \underline{0.0353} & $\mathbf{0.0028}$
& 0.0975 & 7.2869
& 0.2052 & 10.2803 \\

& \xmark & 20 
& 79.53
& 0.2830 & 0.0053 & 0.2051 & 0.1479 & 0.0601
& 0.0993 & 7.9207
& 0.2692 & 9.2774 \\

\tabspacerow
\hline
\tabspacerow

GD (PyTorch3D)
& \xmark & 10 
& 98.59
& 0.5279 & 0.0108 & 0.4162 & 0.2571 & 0.0501
& 0.1445 & 8.4748
& 0.2306 & 8.8919 \\

\tabspacerow
\hline
\tabspacerow

XNES
& \xmark & 3
& \underline{15.45}
& 0.4565 & 0.0075 & 0.2526 & 0.2977 & 0.1397
& 0.0913 & 6.7764
& 0.1417 & 7.5942 \\

\tabspacerow
\hline
\tabspacerow

\multirow{3}{*}{\centering CMA-ES (Ours)}
& \xmark & 3
& 15.93
& 0.2136 & 0.0032 & 0.1371 & 0.1335 & 0.0705
& 0.0729 & \underline{5.7210}
& 0.1107 & \underline{6.9970} \\

& \bcheckmark & 3 
& 15.85
& $\mathbf{0.0550}$ & $\mathbf{0.0015}$ & $\mathbf{0.0066}$ & $\mathbf{0.0144}$ & \underline{0.0043}
& 0.0757 & 6.1491
& 0.1066 & 7.3529 \\

& \xmark & 5
& 23.90
& \underline{0.0734} & $\mathbf{0.0015}$ & 0.0518 & 0.0535 & 0.0241
& \underline{0.0674} & $\mathbf{5.1007}$
& $\mathbf{0.0931}$ & $\mathbf{6.3607}$ \\

\tabspacerow
\hline
\tabspacerow


\multirow{1}{*}{\centering CMA-ES (Online)}
& \bcheckmark & 3
& $25.15^\ast$
& -- & -- & -- & -- & --
& $\mathbf{0.0662}$ & 6.0133
& \underline{0.1004} & 7.1115 \\

\tabspacerow
\hline
\tabspacerow

\multirow{1}{*}{CMA-ES (w/o kpt. loss)}

& \xmark & 3
& 15.77
& 0.2356 & \underline{0.0028} & 0.1279 & 0.2015 & 0.1444
& 0.0727 & 14.9398
& 0.1020 & 24.5069 \\

\tabspacerow
\hline
\tabspacerow

\multirow{1}{*}{CMA-ES (OpenCV kpt.)}

& \xmark & 3
& $\mathbf{13.63}$
& 0.7706 & 0.0177 & 0.6312 & 0.9356 & 0.7557
& 0.0998 & 9.9159
& 0.1484 & 12.5161 \\

\tabspacerow


\tablesection{Dual-Arm Tracking}
\tabspacerow

\multirow{2}{*}{\centering GD}

& \xmark & 10
& 57.46
& 0.5169 & 0.0093 & 0.3412 & 0.3336 & 0.1897
& 0.1545 & 10.4062
& 0.2739 & 9.7430 \\

& \bcheckmark & 10  
& 57.52
& \underline{0.1943} & \underline{0.0055} & \underline{0.0341} & \underline{0.0351} & $\mathbf{0.0029}$
& \underline{0.0924} & \underline{6.8002}
& 0.1914 & 9.3923 \\

\tabspacerow
\hline
\tabspacerow

\multirow{2}{*}{\centering CMA-ES}

& \xmark & 3 
& \underline{21.69}
& 0.5068 & 0.0090 & 0.2999 & 0.3753 & 0.1959
& 0.1004 & 6.8037
& \underline{0.1625} & \underline{8.1210} \\

& \bcheckmark & 3
& $\mathbf{21.68}$
& $\mathbf{0.1736}$ & $\mathbf{0.0049}$ & $\mathbf{0.0128}$ & $\mathbf{0.0221}$ & \underline{0.0047}
& $\mathbf{0.0832}$ & $\mathbf{6.0193}$
& $\mathbf{0.1268}$ & $\mathbf{7.1138}$ \\

\tabspacerow
\hline

\end{tabular}}
\caption{
Quantitative results and ablations on the synthetic dataset and the SurgPose dataset. Except for the online version, the reference masks are pre-computed using Surgical SAM~2. \textbf{Setup:} ``Joint'' indicates whether joint angle readings are used, and ``Iters'' denotes the number of iterations per frame. Runtimes report the per-frame inference latency (ms) for optimization; for the online version of our method, the runtime marked with an asterisk includes Surgical SAM~2 inference. Rotation and joint angle errors are reported in radians (rad), translation error in meters (m), mask error as $1 - \mathrm{IoU}$, and keypoint reprojection error in pixels. For dual-arm tracking, each arm is evaluated separately, and results are averaged across the two arms. Best results are in \textbf{bold} and second-best are underlined, respectively. 
}

\label{tab:main_comparison}
\vspace{-0.14in}
\end{table*}

\begin{table}[t]
\centering
\begin{tabular}{lcc}
\hline
\textbf{Method} & \textbf{Mask Err.} & \textbf{FPS} \\
\hline
 Richter \textit{et al.} (1k particles) \cite{Florian_filter} & $0.2476 \pm 0.0515$ & $ 19.22 \pm  0.45$ \\
Ours (w/ joint, 3 iter\,/\,frame) & $0.1177 \pm 0.0396$ & $43.34 \pm 1.61$ \\
\hline
\end{tabular}

\caption{Comparison of online tracking accuracy and inference speed between Richter \textit{et al.} \cite{Florian_filter} and the proposed method. All values are reported as $\text{mean} \pm \text{std}$.}
\label{tab:pf_comparison}
\vspace{-0.14in}
\end{table}
 
Table~\ref{tab:main_comparison} reports the quantitative results on the synthetic dataset and the SurgPose dataset. The proposed framework based on CMA-ES is compared against gradient descent (GD) baselines and XNES \cite{XNES}, a natural evolution strategy closely related to CMA-ES. In gradient-based ablations, NvDiffRast is replaced with PyTorch3D \cite{ravi2020pytorch3d}, as used in prior works \cite{DR_orig, liang2025differentiablerenderingbasedposeestimation}.

For the synthetic dataset with ground-truth kinematics data, pose reconstruction accuracy is evaluated in terms of rotation, translation, and joint angle errors. Rotation error is computed as $\|\log(\mathbf{R}^\top \hat{\mathbf{R}})^\vee\|_2$, where $\mathbf{R}$ and $\hat{\mathbf{R}}$ denote the ground-truth and estimated orientations, respectively. Translation error is measured as the $\ell_2$ distance, and absolute errors are reported for the three visible joints: wrist pitch $q_1$, wrist yaw $q_2$, and jaw angle $q_3$. To resolve the pose ambiguity of LND, the pose with the smaller absolute error in the local $z$-axis rotation $\beta$ is selected among the two equivalent solutions for evaluation. 

For the SurgPose dataset, raw joint angle readings require compensation and cannot serve as direct ground truth. Therefore, the Intersection-over-Union (IoU) between the reference mask $\mathbb{M}_\text{ref}$ from Surgical SAM~2 and the rendered mask $\mathbb{S}$ from the pose estimation is computed, and the mask error is defined as $1 - \text{IoU}(\mathbb{M}_\text{ref}, \mathbb{S})$. When ground-truth reference keypoints $\mathbf{t}^r$ are available, the reprojection error is computed as the sum of $\ell_2$ distances between corresponding keypoints. For fair comparison, we use pre-computed reference masks $\mathbb{M}_\text{ref}$ for all methods. 
To demonstrate end-to-end performance, we additionally include a fully online variant of our approach, where segmentation masks are generated by Surgical SAM~2 on the fly.

Across both single- and dual-arm tracking, CMA-ES achieves the best balance between accuracy and efficiency. With only 3 iterations per frame, it consistently outperforms GD with 10--20 iterations in pose reconstruction for both the synthetic and real-world data while requiring approximately $37\%$ of the runtime. Incorporating joint angle readings further improves performance and yields the best overall results. Although XNES is more efficient than GD, it remains inferior to CMA-ES in reconstruction accuracy, as the CMA-ES update rule (\ref{eqn:cmaes_update}) enables faster convergence to local quadratic optima \cite{XNES}. Substituting NvDiffRast with PyTorch3D in the gradient-based setting results in both reduced accuracy and increased runtime. Furthermore, the contribution of the keypoint loss $\mathcal{L}_{\text{kpts}}$ in (\ref{eqn:kpts_loss}) is assessed through two ablations: (1) removing the keypoint term slightly degrades performance, while (2) replacing the trained keypoint detection model with the traditional GoodFeaturesToTrack method \cite{opencv} yields the worst overall performance, underscoring the importance of robust keypoint supervision.

Table~\ref{tab:pf_comparison} reports the quantitative results of benchmarking the proposed method against the particle filters proposed by Richter \textit{et al.} \cite{Florian_filter} for online tool tracking on the collected dataset. The mask error $(1 - \text{IoU})$ and inference speed (FPS) are reported for both methods. For the proposed method, the reported FPS accounts for Surgical SAM~2 prediction, keypoint detection, and CMA-ES optimization. The method demonstrates clear improvements in both accuracy and efficiency, benefiting from a GPU-parallelized implementation. 

\section{Discussions and Conclusions}

In this work, we present a robust and generalizable real-time surgical instrument tracking framework based on evolutionary optimization and batched rendering. Through extensive qualitative and quantitative evaluations on both synthetic and real-world datasets, we demonstrate that the proposed approach effectively addresses key limitations of prior methods, including noisy joint angle readings, suboptimal convergence, and long iterative optimization. These improvements in accuracy and runtime performance enhance its practicality for downstream clinical and robotic applications.

A remaining limitation of rendering-based methods lies in their dependence on accurate segmentation. Errors caused by occlusion or noisy segmentation propagate directly to pose and kinematic estimation. Furthermore, the SurgPose dataset does not contain sequences with overlapping instruments, preventing evaluation of our bi-manual tracking method under frequent instrument–instrument occlusions encountered in real surgical procedures. In future work, we will incorporate more robust instrument-level segmentation strategies to better handle occlusion scenarios. We also plan to leverage high-throughput batch rendering platforms (e.g., \cite{10.1145/3680528.3687629}) to fully exploit the blackbox optimization paradigm of evolutionary strategies and further improve runtime efficiency.

\bibliography{IEEEabrv, reference}

\end{document}